\documentclass[10pt,twocolumn,letterpaper]{article}

\usepackage[pagenumbers]{cvpr}              %
\usepackage{amsmath}
\usepackage{multirow}
\usepackage{booktabs}
\usepackage{colortbl}
\usepackage{siunitx}
\usepackage{algorithm}
\usepackage{algorithmic}

\newcommand{\myparagraph}[1]{\vspace{1pt}\noindent{\bf{#1}}}

\definecolor{DnCBG}{rgb}{0.9, 0.9, 1.}

\definecolor{cvprblue}{rgb}{0.21,0.49,0.74}
\usepackage[pagebackref,breaklinks,colorlinks,allcolors=cvprblue]{hyperref}

\def\paperID{26} %
\def\confName{CVPR}
\def\confYear{2025}

\title{A Good \texttt{CREPE} needs more than just \texttt{Sugar}:\\ Investigating Biases in Compositional Vision-Language Benchmarks}

\author{
  Vishaal Udandarao$^{1,3*}$
  \qquad Mehdi Cherti$^{2*}$ \qquad Shyamgopal Karthik$^{1}$ \\
  \qquad Jenia Jitsev$^{2}$ \qquad Samuel Albanie$^{\dagger}$ \qquad Matthias Bethge$^{1\dagger}$
~\\
\small{
$^1$University of Tübingen and Tübingen AI Center \quad
$^2$Juelich Supercomputing Centre (JSC)\quad
$^3$University of Cambridge \quad
}\\
 }
\begin{document}

\maketitle

\begin{abstract}
We investigate $17$ benchmarks (\textit{e.g.}, SugarCREPE, VALSE) commonly used for measuring compositional understanding capabilities of vision-language models (VLMs). We scrutinize design choices in their construction, including data source (\textit{e.g.}, MS-COCO) and curation procedures (\textit{e.g.}, constructing negative images/captions), uncovering several inherent biases across most 
benchmarks. We find that blind heuristics (\textit{e.g.}, token-length, log-likelihood under a language model) perform on par with CLIP models, indicating that these benchmarks do not effectively measure compositional understanding. We demonstrate that the underlying factor is a \textit{distribution asymmetry} between positive and negative images/captions, induced by the benchmark construction procedures. 
To mitigate these issues, we provide a few key recommendations for constructing more robust vision-language compositional understanding benchmarks, that would be less prone to such simple attacks.

\end{abstract}

\section{Introduction}
\label{sec:intro}
Compositionality is the notion that the ``\textit{the meaning
of the whole is a function of the meanings of its parts}''~\cite{cresswell2016logics}. %
Several prior works~\cite{stone2017teaching,lake2017building,hudson2018compositional,grunde2021agqa} posit that visio-linguistic compositional reasoning is essential. Consequently, several benchmarks have been proposed to evaluate the compositional understanding abilities of VLMs~\cite{sugarcrepe,crepe,aro,winoground,vlchecklist,valse,evilprobe,kamath2024hard,tong2024eyes,zeng2024investigating,kamath2023text,awal2024vismin,bevnova2024cv}. These benchmarks involve either (1) matching an image to its correct caption from a set of similar captions (image-to-text retrieval), or  (2) matching an input text caption to its correct image from a set of similar images (text-to-image retrieval), or (3) performing both these tasks (matching a set of images and texts correctly to each other). While VLMs initially performed almost at chance level~\cite{winoground,aro}, recent works have been able to significantly improve their performance~\cite{mitra2023compositional,zhang2023contrasting,castro2024clove,patel2024tripletclip,wang2024enhancing,oh2024preserving,li2025enhancing,koishigarina2025clip,zheng2024iterated}.  

\begin{figure}[!t]
    \centering
    \includegraphics[width=\linewidth]{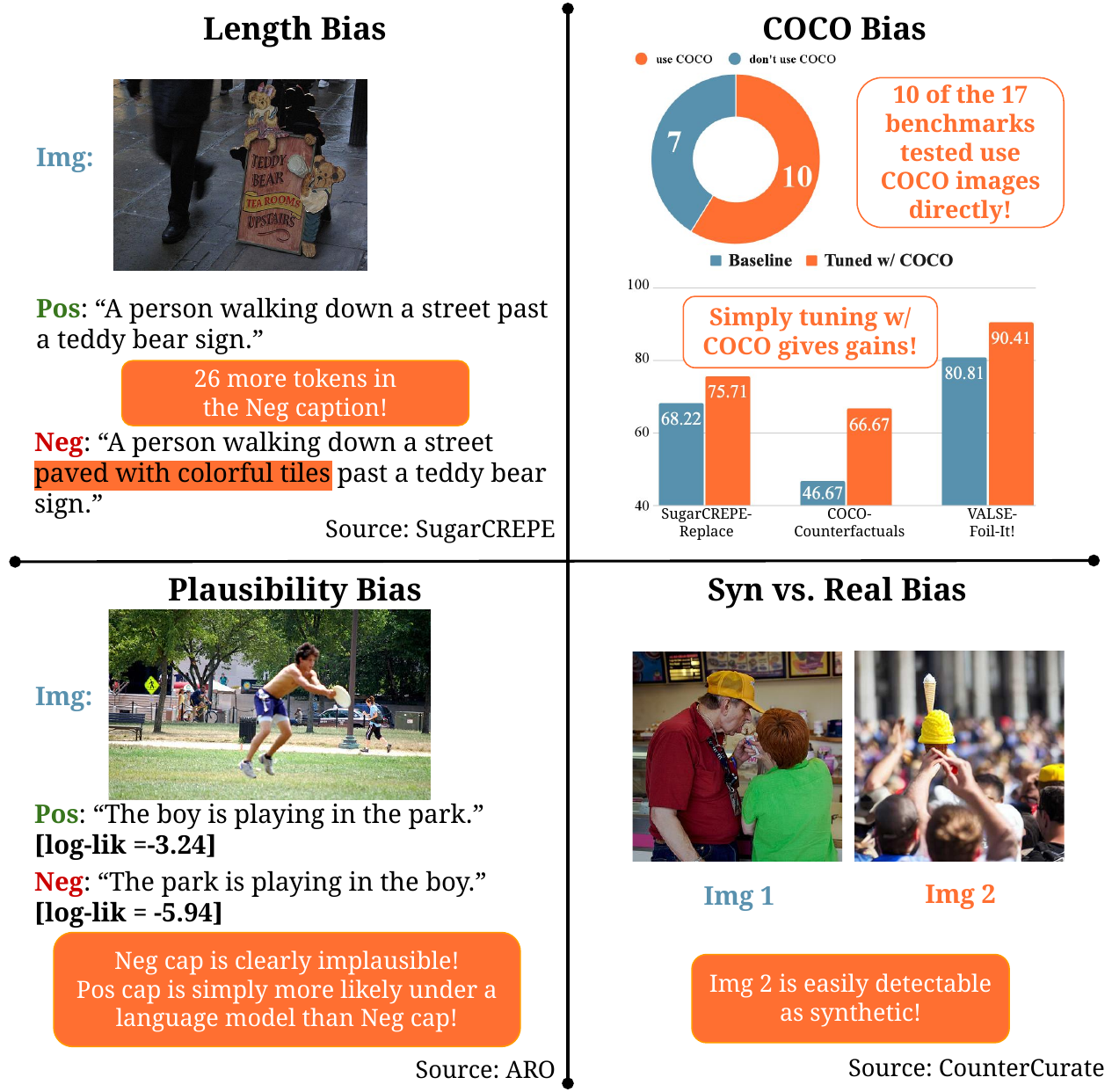}
    \caption{\textbf{Biases in Compositionality Benchmarks.}
    We identify $4$ biases, arising from data sources (\textit{e.g.}, COCO bias) and the construction of negative samples (\textit{e.g.}, length bias), creating a \textit{distributional asymmetry} between positives and negatives. This allows simple unimodal heuristics like token-length or LLM likelihood to match CLIP performance, questioning these benchmarks' validity.}
    \label{fig:teaser}
    \vspace{-15pt}
\end{figure}

\begin{table*}[!t]
\hspace*{-1cm}
\scriptsize
\centering
\setlength{\tabcolsep}{3pt} %
\renewcommand{\arraystretch}{1.2} %
\caption{\textbf{Overview of Current Vision-Language Compositionality Benchmarks.} We present statistics of $17$ datasets, including their primary data sources, the tasks they assess, and the compositional primitives they operate on. The \textbf{Primitives} column denotes the axes of compositionality: attributes (\texttt{Att}), relations (\texttt{Rel}), objects (\texttt{Obj}), and ordering (\texttt{Ord}). For \textbf{Text Types}, \texttt{R} indicates real captions from human annotators or web alt-texts, \texttt{T} denotes templated captions from handcrafted templates, and \texttt{S} represents those generated or edited by a large language model (LLM). For \textbf{Image Types}, \texttt{R} refers to naturally collected images, while \texttt{S} indicates those generated by a text-to-image diffusion model. ${\dagger}$ We consider the \texttt{productivity} subset as it is applicable to all VLMs, regardless of pretraining datasets.}
\begin{tabular}{lrcccccc}
\toprule
\textbf{Benchmark} & \textbf{Size} & \textbf{Data Source} & \textbf{Task} & \textbf{Primitives} & \textbf{Image Types} & \textbf{Text Types} & \textbf{License} \\
\midrule
ARO \cite{aro} & 58,685 & VG, GQA, COCO, Flickr & I$\rightarrow$T & \texttt{Att,Rel,Ord} & \texttt{R} & \texttt{R/T} & MIT \\
CREPE$^{\dagger}$ \cite{crepe} & 42,406 & VG & I$\rightarrow$T & \texttt{Att,Rel,Obj} & \texttt{R} & \texttt{S/T} & Unknown \\
SugarCREPE \cite{sugarcrepe} & 7,512 & COCO & I$\rightarrow$T & \texttt{Att,Rel,Obj} & \texttt{R} & \texttt{R/S} & MIT \\
VL-Checklist \cite{vlchecklist} & 46,833 & VG, SWiG, HAKE, VAW & I$\rightarrow$T & \texttt{Att,Rel,Obj} & \texttt{R} & \texttt{R/T} & Unknown \\
VALSE \cite{valse} & 8,782 & Visual7W, COCO, SWiG, VisDial & I$\rightarrow$T & \texttt{Att,Rel,Obj} & \texttt{R} & \texttt{R/T/S} & MIT \\
ColorFoil \cite{colorfoil} & 3,017 & COCO, Flickr & I$\rightarrow$T & \texttt{Att} & \texttt{R} & \texttt{R/T} & MIT \\
BLA \cite{bla} & 1,939 & VG & I$\rightarrow$T & \texttt{Att} & \texttt{R} & \texttt{T} & MIT \\
What's-Up \cite{whatsup} & 4,958 & COCO, GQA & I$\rightarrow$T & \texttt{Rel} & \texttt{R} & \texttt{R/T} & MIT \\
CC-Neg \cite{singh2024learn} & 228,246 & CC-3M & I$\rightarrow$T & \texttt{Att,Rel,Obj} & \texttt{R} & \texttt{R/S} & MIT \\
SugarCREPE++ \cite{sugarcrepe++} & 4,759 & COCO & I$\rightarrow$T & \texttt{Att,Rel,Obj} & \texttt{R} & \texttt{R/S} &  CC-BY-4.0 \\
\midrule
SVO-Probes \cite{svo} & 36,841 & Google Images & T$\rightarrow$I & \texttt{Rel,Obj} & \texttt{R} & \texttt{R} & Apache-2.0 \\
COLA \cite{cola} & 32,371 & VG, GQA, PACO, CLEVR & T$\rightarrow$I & \texttt{Att,Obj} & \texttt{R} & \texttt{R/T} & MIT \\
\midrule
Winoground \cite{winoground} & 400 & Getty Images & Both & \texttt{Att,Rel,Obj,Ord} & \texttt{R} & \texttt{R} & Custom \\
ColorSwap \cite{colorswap} & 300 & Text-to-Image Models & Both & \texttt{Att} & \texttt{S} & \texttt{R/T/S} & MIT \\
COCO-Counterfactuals \cite{cococounterfactuals} & 34,820 & Text-to-Image Models & Both & \texttt{Obj} & \texttt{S} & \texttt{R/S} & CC-BY-4.0 \\
EqBen-Video \cite{wang2023equivariant} & 243,099 & YC2, AG, GEBC & Both & \texttt{Att,Rel,Obj} & \texttt{R} & \texttt{R/T} & MIT \\
CounterCurate \cite{countercurate} & 11,343 & Flickr & Both & \texttt{Rel} & \texttt{R/S} & \texttt{R/S} & CC \\
\bottomrule
\end{tabular}
\label{tab:benchmark-comparison}
\end{table*}

\noindent Given this increased attention, it is crucial to audit current benchmarks to ensure they accurately measure the desired properties, rather than being solvable through unintended shortcuts~\cite{geirhos2020shortcut}. We ask---\textit{do current vision-language compositionality benchmarks contain biases that offer ``shortcuts'' to VLMs? If so, what are these biases, why do they arise, and how can we mitigate them?} Although prior work has briefly explored this~\cite{sugarcrepe,li2024removing,evilprobe,cappa}, we comprehensively uncover several concrete issues, identify their root causes grounded in current curation procedures, and provide some simple recommendations to mitigate these issues.

\noindent To concretely answer these questions, we survey a set of $17$ popular benchmarks that are used for evaluating VLMs (see~\cref{tab:benchmark-comparison}). 
These benchmarks vary widely in their data-sources, task-types, and negative text/image construction strategies. 
However, these different design choices can introduce several biases, potentially compromising the benchmark quality.
Most of these benchmarks generate ``hard-negatives'' using handcrafted rules~\cite{aro,crepe}, large language models (LLMs)~\cite{sugarcrepe,valse}, or text-to-image models~\cite{countercurate,colorswap}, introducing simple shortcuts for solving the task.
For example, in the ARO benchmark~\cite{aro}, an image, whose positive caption is ``The boy is playing in the park'', has a constructed negative caption as ``The park is playing in the boy.'' (see~\cref{fig:teaser} bottom left), which is obviously implausible.

\noindent We systematically uncover $4$ such biases across all these benchmarks (see~\cref{fig:teaser}). We show that blind models (\textit{e.g.}, text-only LLMs) match performance of CLIP on several  image-to-text benchmarks, finding that the hard negatives can be easily distinguished from positives due to systemic differences in token-length or caption-likelihoods. Additionally, in text-to-image benchmarks, where hard-negative images are generated using text-to-image models, one can easily distinguish the negative images from real positive images (see~\cref{fig:teaser} bottom right). We identify a \textit{distributional asymmetry} between the properties of positive and negative images/captions to be the key underlying factor for these biases. This asymmetry arises due to the lack of careful curation during the negative sample construction phase.

\noindent Finally, we provide a set of recommendations to improve the state of visio-linguistic compositional evaluation. Firstly, merely filtering out samples with distributional asymmetry is insufficient since there will always exist another model that captures new asymmetries. For instance, SugarCREPE was curated using grammar~\citep{morris2020textattack} and common-sense models~\cite{vera}, however different models (even from \textit{the same family}) still exploit asymmetries between the positive and negative captions. Hence, meaningful solutions to this problem could involve having additional negatives in addition to addressing basic distributional asymmetry issues.
Finally, evaluations that simultaneously match image and text together further reduce the likelihood of benchmark-hacking by blind baselines.

\section{The Issue of \textit{Distributional Asymmetry}}

\myparagraph{Benchmarks.} As a starting point, we survey the field of vision-language compositionality, and investigate $17$ commonly used benchmarks. We characterize these benchmarks in~\cref{tab:benchmark-comparison} via their data sources, task-types, and their data curation procedures. For example, SugarCREPE~\citep{sugarcrepe} is mainly sourced from the COCO~\citep{coco} dataset and is constructed as a two-way image-to-text retrieval problem---for a given COCO image, the model is tasked to correctly identify the ``positive'' caption that matches the image over a synthetically constructed ``negative'' caption.

\myparagraph{Simple heuristics as baselines.} We first consider a ``blind'' version of the compositionality tasks, where we feed a language model only the positive and negative captions across the different compositionality benchmarks, ignoring the images as inputs. This helps us understand whether these benchmarks can be solved without any visual inputs, following VQA literature~\citep{goyal2017making,anand2018blindfold}. 
We use LLaMA2-13B~\citep{llama2} as our first ``blind-baseline''. For each pair of positive and negative captions, we compute the log-likelihood under the language model. Consider the positive caption as $C_p = \{ p_1, p_2, \ldots, p_l \}$ and the negative caption as $C_n = \{ n_1, n_2, \ldots, n_q \}$, where \( p_i \) are the positive tokens, \( l \) is the length of the positive caption, \( n_i \) are the negative tokens, and \( q \) is the length of the negative caption. We consider two versions of log-likelihoods, one normalized by each caption's token length and the other unnormalized:
\begin{multline*}
\log P(C) = \sum_{i=1}^{|C|} \log P(c_i \mid c_1, \ldots, c_{i-1})\\
L_{\text{unnorm}}(C_p) = \log P(C_p), L_{\text{unnorm}}(C_n) = \log P(C_n)\\
L_{\text{norm}}(C_p) = \frac{1}{l}\log P(C_p), L_{\text{norm}}(C_n) = \frac{1}{q}\log P(C_n)
\end{multline*}
\noindent where $C=\{c_1, c_2, \dots, c_{|C|}\}$ represents a caption token sequence and \( P(c_i \mid c_1, \ldots, c_{i-1}) \) are the computed token probabilities.
For a given ($C_p$, $C_n$) test sample, we predict the caption with the higher likelihood as correct.
The accuracy obtained by using $L_{\text{norm}}$ represents \textit{plausibility bias}, as it directly measures how likely the positive and negative captions are, in relation to each other, when ignoring the image input. 
The $L_{\text{unnorm}}$-accuracy accounts for the \textit{length bias}, since it factors in the token-length of the captions.

\myparagraph{A Blind VLM baseline.} As a second baseline, we train Cap-models~\citep{tschannen2024image}, using a ViT-B-32 vision-encoder and text-decoder with the same shape as the encoder, except half the depth. We train on DataComp-1B~\citep{datacomp} for 1.28B samples seen with $16$k batch-size. We train two variants---(1) with image-text captions as default (BlindCap (I+T)), and (2) explicitly only on texts. 
We train BlindCap (I+T) by either feeding image-caption pairs or captions only (where only text-decoder is used), to support both image captioning and text generation with the same model. To train BlindCap (T), we simply train the text-decoder using captions only, ignoring the images.
We also further fine-tune these models on the COCO dataset~\citep{coco} to get two variants (BlindCap \textcolor{darkgray}{\scriptsize{COCO-FT}}).
At test time, we only feed in the positive and negative captions, intentionally making these models ``blind''.
For predicting the correct caption, we use normalized log-likelihoods ($L_\text{norm}$).
As before, the goal of this test is to characterize \textit{plausibility bias} under a blind-VLM. Further, note that the accuracy gains yielded by COCO-fine-tuning represents the \textit{COCO bias} inherent in these benchmarks.

\myparagraph{Distributional Asymmetry.} As a first step, we run our LLaMA baselines on $12$ benchmarks. We aim to characterize how different the distributions of the positive and negative captions in these benchmarks are, using likelihood scoring. In particular, we compute the likelihood difference:

\[
\text{Lik-Diff}{=}{\frac{1}{|B|}\sum_{i=1}^{B} \left( L_{\text{unnorm}}(C_p^i) - L_{\text{unnorm}}(C_n^i) \right)}
\]

\noindent where $|B|$ represents the number of samples in a given target benchmark, $C_p^i$ represents the positive caption of the $i^{\text{th}}$ sample and $C_n^i$, the corresponding negative caption.

\begin{figure}
    \centering
    \includegraphics[width=\linewidth]{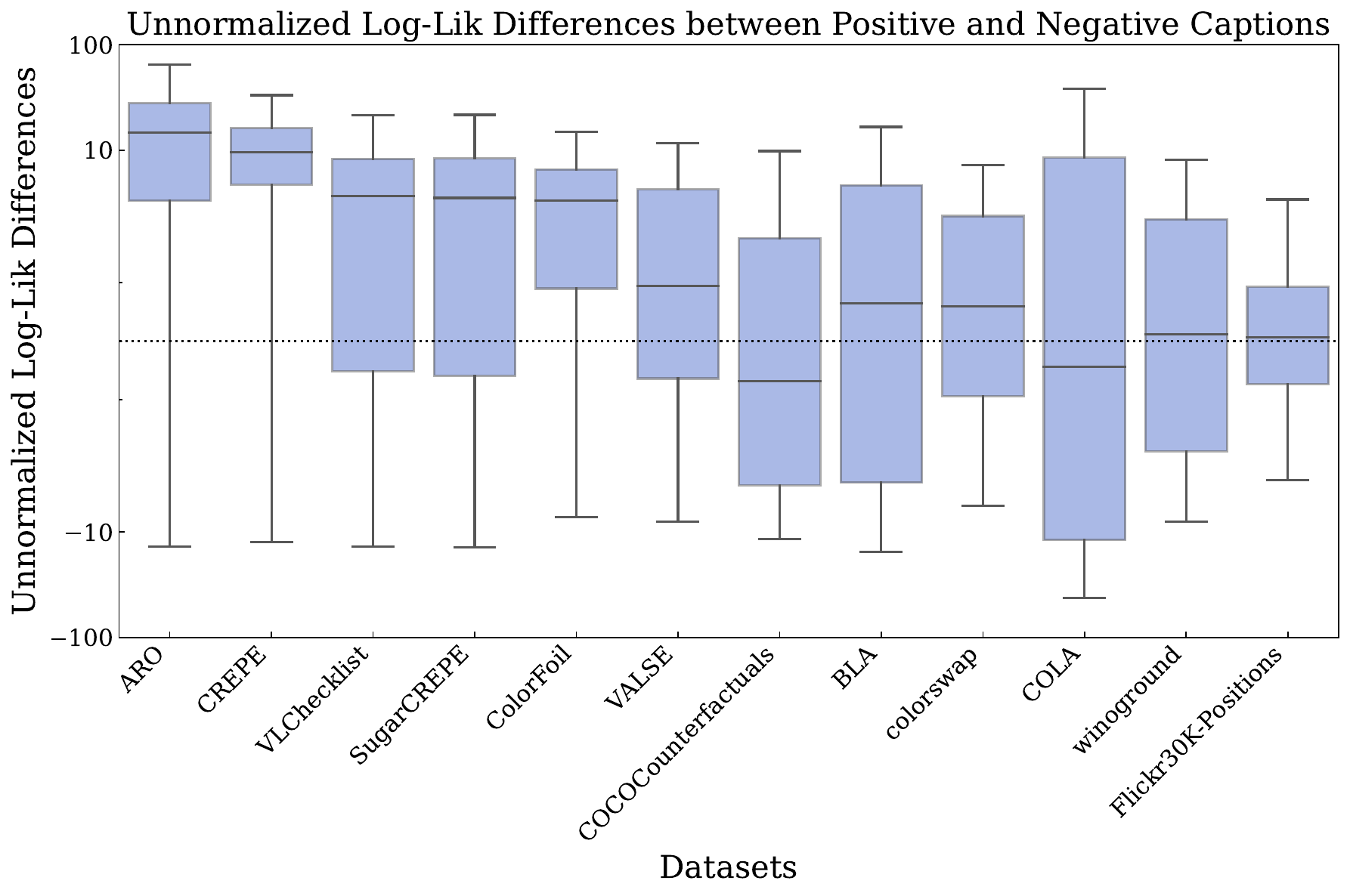}
    \caption{\textbf{Positives and Negatives are distinguished by Blind Likelihood Scoring.} We observe that on several benchmarks (ARO, VL-CheckList, SugarCREPE etc.), the difference in likelihoods between the positive and negative captions with an LLM~\cite{llama2} is itself sufficient to achieve strong performance.}
    \label{fig:unnorm-pos-neg-caption-log-lik-differences}
    \vspace{-10pt}
\end{figure}

\begin{table*}[h]
\footnotesize
\centering
\caption{\textbf{Blind Results on SugarCREPE.} Both LLaMA~\cite{llama2} and our BlindCap~\cite{cappa} models achieve strong results without any image inputs. Fine-tuning BlindCap on COCO~\cite{coco} further improves performance highlighting the ease of exploiting the COCO-bias.}
\begin{tabular}{lcccccccc}
\toprule
\textbf{Model} & \textbf{Add-Att} & \textbf{Add-Obj} & \textbf{Replace-Att} & \textbf{Replace-Rel} & \textbf{Replace-Obj} & \textbf{Swap-Att} & \textbf{Swap-Obj} & \textbf{Average} \\
\midrule
\textcolor{gray}{Random} & \textcolor{gray}{50.0} & \textcolor{gray}{50.0} & \textcolor{gray}{50.0} & \textcolor{gray}{50.0} & \textcolor{gray}{50.0} & \textcolor{gray}{50.0} & \textcolor{gray}{50.0} & \textcolor{gray}{50.0} \\
\textcolor{gray}{CLIP} & \textcolor{gray}{68.2} & \textcolor{gray}{77.2} & \textcolor{gray}{80.0} & \textcolor{gray}{69.2} & \textcolor{gray}{90.9} & \textcolor{gray}{64.0} & \textcolor{gray}{61.4} & \textcolor{gray}{72.9} \\
\midrule
LLaMA-13B (norm) & 51.6 & 16.3 & 55.3 & 63.6 & 43.2 & 71.5 & 65.9 & 52.5 \\
LLaMA-13B (unnorm) & 93.6 & 92.0 & 59.0 & 67.4 & 46.3 & 69.5 & 68.7 & 70.9 \\
BlindCap (T) & 98.8 & 97.5 & 70.8 & 74.2 & 58.2 & 76.7 & 75.2 & 78.8 \\
BlindCap (I+T) & 99.1 & 97.5 & 70.8 & 75.6 & 58.5 & 77.6 & 74.0 & 79.0 \\
BlindCap (T) \tiny{\textcolor{darkgray}{COCO-FT}} & 99.2 & 98.1 & 74.8 & 81.7 & 71.1 & 78.8 & 72.4 & 82.3 \\
\rowcolor{DnCBG}BlindCap (I+T) \tiny{\textcolor{darkgray}{COCO-FT}} & 99.3 & 97.8 & 74.0 & 81.6 & 71.6 & 78.4 & 74.4 & \textbf{82.4} \\

\bottomrule
\end{tabular}
\label{tab:blind-sugarcrepe-results}
\vspace{-5pt}
\end{table*}

\begin{table*}[h]
\setlength{\tabcolsep}{2pt} %
\centering
\footnotesize
\caption{\textbf{Blind Results on VALSE.} Our models outperform CLIP on 5/10 categories, highlighting the distributional asymmetry issue.}
\begin{tabular}{lcccccccccccc}
\toprule
\textbf{Model} & \textbf{Act-Swap} & \textbf{Act-Replace} & \textbf{Coref-Hard} & \textbf{Coref-Std} & \textbf{Cnt-Adv} & \textbf{Cnt-Hard} & \textbf{Cnt-Small} & \textbf{Exist} & \textbf{Foil-It} & \textbf{Plural} & \textbf{Rel} & \textbf{Average} \\
\midrule
\textcolor{gray}{Random} & \textcolor{gray}{50.0} & \textcolor{gray}{50.0} & \textcolor{gray}{50.0} & \textcolor{gray}{50.0} & \textcolor{gray}{50.0} & \textcolor{gray}{50.0} & \textcolor{gray}{50.0} & \textcolor{gray}{50.0} & \textcolor{gray}{50.0} & \textcolor{gray}{50.0} & \textcolor{gray}{50.0} & \textcolor{gray}{50.0} \\
\textcolor{gray}{CLIP} & \textcolor{gray}{68.6} & \textcolor{gray}{75.6} & \textcolor{gray}{49.7} & \textcolor{gray}{52.1} & \textcolor{gray}{57.5} & \textcolor{gray}{62.1} & \textcolor{gray}{62.5} & \textcolor{gray}{66.9} & \textcolor{gray}{88.8} & \textcolor{gray}{56.2} & \textcolor{gray}{64.3} & \textcolor{gray}{64.0} \\
\midrule
\rowcolor{DnCBG}LLaMA-13B (norm) & 85.1 & 69.6 & 65.3 & 58.0 & 68.0 & 52.0 & 48.6 & 62.0 & 85.3 & 50.2 & 75.6 & \textbf{65.4} \\
LLaMA-13B (unnorm) & 88.6 & 67.1 & 65.3 & 58.0 & 62.7 & 51.9 & 50.0 & 49.4 & 80.8 & 52.6 & 75.7 & 63.8 \\
BlindCap (T) & 86.4 & 60.5 & 51.1 & 47.2 & 59.5 & 49.0 & 49.9 & 49.6 & 81.0 & 54.0 & 75.1 & 60.3 \\
BlindCap (I+T) & 88.1 & 61.5 & 47.5 & 45.0 & 62.7 & 45.9 & 48.4 & 59.0 & 80.8 & 54.9 & 74.4 & 60.7 \\
BlindCap (T) \tiny{\textcolor{darkgray}{COCO-FT}} & 77.1 & 55.1 & 61.0 & 54.5 & 55.0 & 53.1 & 50.4 & 54.9 & 90.9 & 56.2 & 83.1 & 62.8 \\
BlindCap (I+T) \tiny{\textcolor{darkgray}{COCO-FT}} & 80.9 & 56.5 & 53.9 & 52.1 & 56.4 & 48.1 & 51.0 & 60.5 & 90.4 & 55.3 & 82.7 & 62.5 \\
\bottomrule
\end{tabular}
\label{tab:blind-valse-results}
\vspace{-15pt}
\end{table*}

\begin{table}[h]
\footnotesize
\centering
\caption{\textbf{Blind Results on VL-Checklist.} Likelihoods from our blind models achieve non-trivial accuracies on all tasks while matching the performance of CLIP on attributes and relations.}
\begin{tabular}{lccccccccccc}
\toprule
\textbf{Model} & \textbf{Attribute} & \textbf{Object} & \textbf{Relation} & \textbf{Average} \\
\midrule
\textcolor{gray}{Random} & \textcolor{gray}{50.0} & \textcolor{gray}{50.0} & \textcolor{gray}{50.0} & \textcolor{gray}{50.0} \\
\textcolor{gray}{CLIP} & {67.9} & \textcolor{gray}{82.8} & \textcolor{gray}{64.1} & \textcolor{gray}{71.6} \\
\midrule
LLaMA-13B (norm) & 48.7 & 57.9 & 68.4 & 58.3 \\
LLaMA-13B (unnorm) & 68.7 & 73.7 & 70.9 & 71.1 \\
BlindCap (T) & 76.1 & 73.1 & 88.1 & 79.1 \\
\rowcolor{DnCBG}BlindCap (I+T) & 76.9 & 72.8 & 87.8 & \textbf{79.2} \\
BlindCap (T) \tiny{\textcolor{darkgray}{COCO-FT}} & 78.0 & 70.8 & 85.9 & 78.2 \\
BlindCap (I+T) \tiny{\textcolor{darkgray}{COCO-FT}} & 77.0 & 72.3 & 85.7 & 78.3 \\
\bottomrule
\end{tabular}
\label{tab:blind-vlchecklist-results}
\vspace{-15pt}
\end{table}

\noindent If there were to be a symmetric distribution of likelihoods, we would expect $\text{Lik-Diff}$s across benchmarks to be close to $0$. However, from~\cref{fig:unnorm-pos-neg-caption-log-lik-differences} we note the extremely high non-zero likelihood-differences across several benchmarks, highlighting the severe \textit{distribution asymmetry issue}. In particular, ARO, CREPE, VL-Checklist, VALSE, SugarCREPE and ColorFoil have the highest absolute asymmetry. This suggests that these benchmarks can indeed be solved by simple unimodal heuristics, which we explore next.

\myparagraph{Breaking Bad Benchmarks.} We show results of our blind models (LLaMAs and BlindCaps) on SugarCREPE (\cref{tab:blind-sugarcrepe-results}), VALSE (\cref{tab:blind-valse-results}) and VL-Checklist (\cref{tab:blind-vlchecklist-results}). We report chance-level and OpenAI-CLIP-B/32~\citep{clip} performance as reference numbers. For SugarCREPE, we observe that on the `add' subset, unnormalized LLaMA scores are quite strong, suggesting an extreme \textit{length bias} issue.
Additionally, our base BlindCaps nearly achieve perfect scores highlighting the inherent \textit{plausibility bias}. 
The fine-tuned variants further boost our numbers, achieving close to $10{\%}$ gain on average over CLIP.
We also demonstrate similar trends on both VALSE~\cite{valse} and VL-Checklist~\cite{vlchecklist}. 
On majority of the tasks, our blind baselines perform on-par or better than CLIP--- 
significantly, atleast one of our blind baselines outperforms CLIP on average by $1.4\%$ (VALSE), $7.6$\% (VL-Checklist) and $9.5\%$ (SugarCREPE).

\section{Recommendations for future benchmarking 
}
\label{sec:recommendations}

To enhance benchmark robustness, we recommend:

\myparagraph{Sourcing Images and Captions from Same Distribution.} Currently, real captions are used as positives, while negatives are synthetically generated using language or diffusion models, leading to the \textit{distribution asymmetry} issue. We recommend 
sourcing from a single distribution, such as relying exclusively on real or uniformly generated data.

\myparagraph{Utilizing Multiple Positives/Negatives.} Several benchmarks include only one positive-negative caption-pair per image, resulting in $50\%$ chance performance alone. Such setups enable simple heuristics to perform competitively, diminishing the effectiveness of these benchmarks. We suggest incorporating multiple, challenging positive/negative captions~\cite{sugarcrepe++,kamath2024hard} to create more rigorous frameworks.

\setlength\tabcolsep{0.8pt}
\begin{table}[h]
\footnotesize
\centering
\caption{\textbf{Model-based filtering is insufficient.} Other text-only models (even from \textit{the same \texttt{text-attack}~\citep{morris2020textattack} family} (\textcolor{gray}{same-fam}) as the SugarCREPE-filtering-model (\textcolor{gray}{sc-filter})) get high accuracies, signifying that filtering is not a robust solution.}
\begin{tabular}{lccc}
\toprule
\textbf{Model} & \textbf{Add-Obj} & \textbf{Replace-Rel} & \textbf{7-Average} \\
\midrule
\texttt{DB-COLA} \scriptsize{\textcolor{gray}{(sc-filter)}} & 46.2 & 53.1 & 52.7 \\
\midrule
\texttt{DB-MRPC} \scriptsize{\textcolor{gray}{(same-fam)}} & 88.5 & 63.6  & 67.8 \\
\texttt{DB-QQP} \scriptsize{\textcolor{gray}{(same-fam)}} & 96.6 & 63.1 & 68.5 \\
\texttt{flan-t5-xl} & 97.2 & 74.0 & 79.5 \\
\bottomrule
\end{tabular}
\label{tab:filtering-is-insufficient}
\vspace{-20pt}
\end{table}

\myparagraph{Bidirectional Image-Text Matching.} Evaluating compositionality solely through single image-caption matching tasks can be limited. Group-based evaluations, where a model must correctly match multiple captions to multiple images simultaneously, present a significantly more challenging task. Therefore, we recommend adopting metrics similar to the Group score introduced by Winoground~\cite{winoground} to more effectively measure compositional understanding.

\myparagraph{Avoiding Over-Reliance on Single-Model Filtering.} Prior attempts to mitigate biases by filtering negatives using a single model~\cite{vera} have proven insufficient (see~\cref{tab:filtering-is-insufficient}). Such filtered benchmarks remain susceptible to biases of the chosen model. We recommend manually curating images and captions from the same distribution, using multiple image and texts simultaneously matched, rather than relying on post-hoc single-model filtering, that is easily hackable.

\section{Conclusion}
In this work, we presented a systematic analysis of biases in vision-language compositionality benchmarks. We find that these benchmarks have issues with \textit{distributional asymmetry}, leading to strong performance of \textit{blind} models that ignore the image input. We provide a set of recommendations, hoping to lead the field towards more meaningful and robust evaluations of visio-linguistic compositionality.

\section*{Acknowledgements}
VU and SGK thank the International Max Planck Research School for Intelligent Systems (IMPRS-IS). VU also thanks the European Laboratory for Learning and Intelligent Systems (ELLIS) PhD program for support. VU is supported by a Google PhD Fellowship in Machine Intelligence. SA is supported by a Newton Trust Grant. MB acknowledges financial support via the Open Philanthropy Foundation funded by the Good Ventures Foundation. MB is a member of the Machine Learning Cluster of Excellence, funded by the Deutsche Forschungsgemeinschaft (DFG, German Research Foundation) under Germany’s Excellence Strategy – EXC number 2064/1 – Project number 390727645. This research utilized compute resources at the Tübingen Machine Learning Cloud, DFG FKZ INST 37/1057-1 FUGG.

{
    \small
    \bibliographystyle{ieeenat_fullname}
    \bibliography{egbib}
}

\end{document}